\documentclass[10pt,twocolumn,letterpaper]{article}
\usepackage[pagenumbers]{cvpr}
\usepackage[utf8]{inputenc}
\usepackage[T1]{fontenc}
\usepackage{url}
\usepackage{booktabs}
\usepackage{amsfonts}
\usepackage{amsmath,amssymb,amsthm}
\usepackage{nicefrac}
\usepackage{microtype}
\usepackage{xcolor}
\usepackage{graphicx}
\usepackage{algorithm}
\usepackage{algpseudocode}
\usepackage{caption}
\usepackage{subcaption}
\usepackage{multirow}
\usepackage{enumitem}
\usepackage{cuted}

\newcommand{\papertitle}{Recurrent Point Transformer via Test-Time Training}

\newcommand{\displayequationstyle}{}
\newcommand{\compactdisplayspacing}{%
  \setlength{\abovedisplayskip}{4pt plus 1pt minus 1pt}%
  \setlength{\belowdisplayskip}{4pt plus 1pt minus 1pt}%
  \setlength{\abovedisplayshortskip}{2pt plus 1pt}%
  \setlength{\belowdisplayshortskip}{2pt plus 1pt}%
}
\newcommand{\compactmethoddisplays}{%
  \setlength{\abovedisplayskip}{2.5pt plus .5pt minus .5pt}%
  \setlength{\belowdisplayskip}{2.5pt plus .5pt minus .5pt}%
  \setlength{\abovedisplayshortskip}{1.5pt plus .5pt minus .5pt}%
  \setlength{\belowdisplayshortskip}{1.5pt plus .5pt minus .5pt}%
  \setlength{\jot}{1pt}%
}
\AddToHook{env/equation/begin}{\compactdisplayspacing\displayequationstyle}
\AddToHook{env/equation*/begin}{\compactdisplayspacing\displayequationstyle}

\definecolor{cvprblue}{rgb}{0.21,0.49,0.74}
\usepackage[breaklinks,colorlinks,allcolors=cvprblue]{hyperref}
\newcommand{\paperauthors}{Wentao Zhou, Xuweiyi Chen, Charles Saunders, Aruni RoyChowdhury, Zezhou Cheng}
\author{Wentao Zhou\textsuperscript{1,2}\quad
        Xuweiyi Chen\textsuperscript{1}\quad
        Charles Saunders\textsuperscript{3}\quad
        Aruni RoyChowdhury\textsuperscript{3}\quad
        Zezhou Cheng\textsuperscript{1}\\[3pt]
  {\normalsize \textsuperscript{1}University of Virginia, Charlottesville\qquad
   \textsuperscript{2}Georgia Institute of Technology\qquad
   \textsuperscript{3}MathWorks}}

\hypersetup{pdftitle={\papertitle},pdfauthor={\paperauthors}}
\title{\papertitle}

\begin{document}

\maketitle
\begin{strip}
  \captionsetup{hypcap=false}
  \centering
  \includegraphics[width=\linewidth]{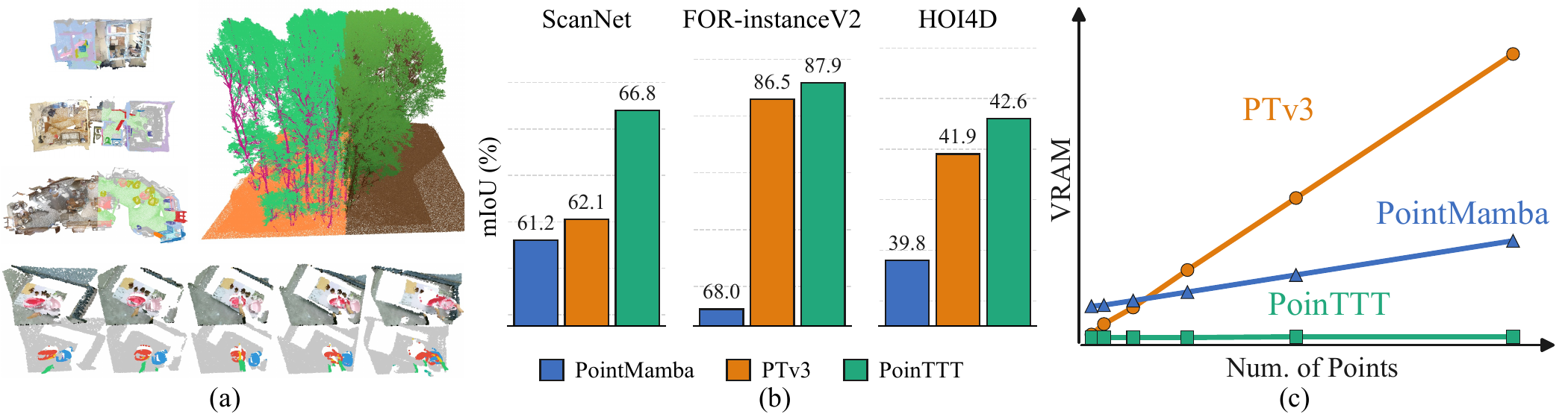}
  \captionof{figure}{\textbf{PoinTTT scales to arbitrary-size point clouds in space and time.} (a) Real-world point clouds vary widely in scale, from room-scale 3D scans to large forest LiDAR surveys and long 4D sequences. (b) PoinTTT processes these inputs recurrently as fixed-size chunks, using test-time training to aggregate long-context information. It outperforms PTv3~\cite{wu2024ptv3} and Mamba-based baselines~\cite{liang2024pointmamba,liu2025pointnet4d} on static and dynamic segmentation benchmarks, including ScanNet~\cite{dai2017scannet}, FOR-instanceV2~\cite{xiang2025forestformer3d}, and HOI4D~\cite{liu2022hoi4d}. The ScanNet bars use full-scene training (Tab.~\ref{tab:pilot}); the HOI4D blue bar denotes P4Mamba (39.8 mIoU). (c) PoinTTT maintains nearly constant memory as input size grows, unlike standard PTv3~\cite{wu2024ptv3}. Although trained with a fixed point budget per chunk, PoinTTT can be applied recurrently to scenes of arbitrary spatial or temporal extent.
  }
  \label{fig:teaser}
\end{strip}

\begin{abstract}
Real-world point clouds are not naturally bounded: 3D scans grow as sensors move through rooms, LiDAR surveys extend across forest plots, and 4D captures can span thousands of frames. Yet most point-cloud encoders are designed for fixed-size inputs, forcing inference-time cropping or tiling that loses long-range context and requires ad-hoc output stitching. We formulate arbitrary-extent point-cloud
segmentation as recurrent sequence-of-chunks processing and introduce
\textit{PoinTTT}, a recurrent point-cloud encoder built on Point Transformer V3 (PTv3). PoinTTT augments PTv3 with residual Large Chunk Test-Time Training (LaCT) layers in both the encoder and decoder. Each LaCT layer maintains compact fast weights as recurrent state: tokens from each chunk update the fast weights, which are then applied to produce chunk features and passed to the next chunk. This design allows PoinTTT to train on fixed-size chunks while processing scenes
of arbitrary spatial or temporal extent, with activation memory bounded by the chunk size. A pilot study on ScanNet and nuScenes shows that recurrent TTT improves segmentation over PTv3 baselines. On FOR-instanceV2, PoinTTT achieves state-of-the-art forest semantic segmentation of $87.9$ mIoU with nearly constant memory. On HOI4D, online full-sequence recurrent inference carries state across clips rather
than resetting at benchmark boundaries, reaching $42.6$ mIoU; a matched full-sequence Mamba control reaches approximately $39.9$. These results suggest that test-time training is a practical primitive for scalable spatial and temporal point-cloud understanding.
\end{abstract}

\section{Introduction}

Modern 3D perception increasingly operates on observations that are unbounded in
space, time, or both. As shown in Fig.~\ref{fig:teaser}a, indoor scans may span
multiple rooms~\citep{armeni20163d,chang2017matterport3d,dai2017scannet,
yeshwanth2023scannetpp}, 4D captures grow as frames arrive~\citep{liu2022hoi4d},
and dense LiDAR surveys can cover broad outdoor plots~\citep{hackel2017semantic3d,
behley2019semantickitti,caesar2020nuscenes,sun2020waymo,varney2020dales,
xiang2025forestformer3d}. Point clouds naturally represent such expanding
observations, yet most point-cloud backbones assume bounded inputs
\citep{qi2017pointnet,qi2017pointnetplusplus,wang2019dgcnn,
thomas2019kpconv,choy20194d,zhao2021pointtransformer,
wu2022pointtransformerv2,wu2024ptv3,wu2024ppt,yue2025litept,yilmaz2026volt,cao2021shapeconv}.
This creates a mismatch for tasks like semantic segmentation: every point requires a
label, but the needed evidence may lie outside the current crop, tile, or clip.
Partitioning large scenes therefore limits context and forces the model to reset
at artificial boundaries, rather than accumulating information across the full
scene or sequence.

We study a different interface for point cloud understanding: rather
than treating cropping or tiling as an external inference procedure, we make
chunking part of the model. An arbitrary-extent scene is represented as a
sequence of fixed-size chunks processed by a recurrent encoder. The model sees
one bounded chunk at a time, keeping training and activation memory controlled,
while carrying a compact state across chunks so inference can scale with the
spatial or temporal extent of the observation. For point clouds, this interface
is practical: the same model can consume indoor scans, forest plots, and 4D
frame streams without a memory footprint that grows with the number of points.

The key design question is where recurrence should be introduced and what form
the recurrent state should take. A strong local backbone such as
PTv3~\citep{wu2024ptv3} already provides effective geometric ordering and
neighborhood reasoning within each chunk, but it has no explicit mechanism for
carrying information across chunks. Recent state-based sequence models,
including linear attention~\citep{katharopoulos2020linear,yang2024parallelizing,team2025kimi},
Mamba~\citep{gu2023mamba,dao2024transformers,lahotimamba}, and Test-time Training~\citep{zhang2025tttright,sun2025ttt}, show
that compact recurrent states can replace explicit all-pair context and improve
efficiency. Our goal is complementary: rather than replacing the point-cloud
backbone, we add a cross-chunk memory module to it. Test-time training provides
a natural mechanism for this memory by storing context in fast weights that are
updated by each chunk and reused by subsequent chunks.

We instantiate this idea as \emph{PoinTTT}, a recurrent point-cloud encoder
built upon PTv3~\citep{wu2024ptv3}. PoinTTT augments both encoder and decoder of PTv3~\citep{wu2024ptv3} with
residual LaCT~\citep{zhang2025tttright} layers. Each LaCT layer maintains a
small fast-weight network as recurrent state. Given a chunk, its tokens update
the fast weights; the updated weights are then applied to produce chunk
features and carried forward to the next chunk. The residual is gated from
zero, so PoinTTT retains PTv3 at initialization and learns to add cross-chunk
computation during training. The same recurrent encoder supports both offline
and online settings: when the whole sequence is available, PoinTTT uses a
bidirectional update-and-apply pass for full-scene spatial inference; for
online 4D input, it propagates the same fast-weight state causally through the
temporal stream, predicting each arriving chunk without future chunks (Sec.~\ref{sec:pointtt}).

We first conduct a pilot study on indoor ScanNet~\citep{dai2017scannet} and
outdoor nuScenes~\citep{caesar2020nuscenes}. On ScanNet, we find that feeding chunks bottlenecks the
network, regardless of whether the network is trained on entire ScanNet scenes
or trained chunk-wise. This supports our hypothesis that partial observation is
suboptimal for point-cloud understanding and highlights the necessity of using
context from the entire scene. We further compare PointMamba~\citep{liang2024pointmamba} and PoinTTT, and
find that PoinTTT outperforms PointMamba by a large margin. This indicates that
test-time training provides stronger performance in this comparison
(see Sec.~\ref{sec:pilot}). On nuScenes, temporal adaptation also improves
PTv3 under single-pass evaluation without test-time augmentation. Furthermore, we demonstrate
PoinTTT as a unified point-cloud understanding backbone for both temporal and
spatial settings. We stress-test PoinTTT on dense tree segmentation using
FOR-instanceV2 against methods tailored for this task, such as
ForestFormer3D~\citep{xiang2025forestformer3d}, and PoinTTT achieves
state-of-the-art performance (see Sec.~\ref{sec:exp_forest}). We then compare online
4D segmentation methods on HOI4D~\citep{liu2022hoi4d} (Sec.~\ref{sec:exp_hoi4d}),
where PoinTTT carries state across complete videos and outperforms a matched
recurrent Mamba baseline.

Our contributions are threefold.
(1) We formulate arbitrary-extent point-cloud segmentation as recurrent chunk processing across spatial and temporal regimes (Sec.~\ref{sec:pilot}, Sec.~\ref{sec:method}, Fig.~\ref{fig:teaser})
(2) We introduce \textit{PoinTTT}, augmenting PTv3~\citep{wu2024ptv3} with residual LaCT~\citep{zhang2025tttright} layers that carry compact fast-weight states for offline full-scene and online temporal inference (Sec.~\ref{sec:method}).
(3) We demonstrate strong performance on ScanNet~\citep{dai2017scannet}, FOR-instanceV2~\citep{xiang2025forestformer3d}, HOI4D~\citep{liu2022hoi4d}, and nuScenes~\citep{caesar2020nuscenes}, with activation memory bounded by chunk size (Sec.~\ref{sec:pilot}, Sec.~\ref{sec:exp_forest}, Sec.~\ref{sec:exp_hoi4d}). Across all settings, PoinTTT processes arbitrary-extent inputs with strong performance and activation memory bounded by chunk size.

\begin{table}[t]
  \centering
  \caption{Pilot results on ScanNet with $10{,}240$-point chunked inference and on nuScenes with single-pass evaluation without TTA. The training-input setting applies to both datasets. PointMamba~\citep{liang2024pointmamba} is a point-cloud state-space baseline; recurrent PoinTTT improves under the same bounded-memory protocol.}
  \label{tab:pilot}
  \label{tab:nuscenes}
  \small
  \setlength{\tabcolsep}{3pt}
  \begin{tabular*}{\linewidth}{@{\extracolsep{\fill}}llcc@{}}
    \toprule
    Backbone & Training input &
    \begin{tabular}[c]{@{}c@{}}ScanNet\\mIoU\end{tabular} &
    \begin{tabular}[c]{@{}c@{}}nuScenes\\mIoU\end{tabular} \\
    \midrule
    \multicolumn{4}{l}{\emph{No recurrent state}} \\
    \midrule
    PTv3~\cite{wu2024ptv3} & full scene & 62.1 & 78.1 \\
    PTv3~\cite{wu2024ptv3} & $10{,}240$ chunks & 63.8 & 77.8 \\
    \midrule
    \multicolumn{4}{l}{\emph{State-space / recurrent state}} \\
    \midrule
    PointMamba~\cite{liang2024pointmamba} & full scene & 61.2 & 73.9 \\
    PoinTTT & full scene & \textbf{66.8} & \textbf{78.7} \\
    \bottomrule
  \end{tabular*}
\end{table}

\section{Pilot Study}
\label{sec:pilot}

We examine segmentation with bounded inputs on indoor
ScanNet~\citep{dai2017scannet} and outdoor
nuScenes~\citep{caesar2020nuscenes}. On ScanNet, all methods in
Tab.~\ref{tab:pilot} use $10{,}240$-point inference chunks, with predictions
reassembled before computing scene-level mIoU. On nuScenes, evaluation is
single-pass without test-time augmentation (TTA). The table's
\emph{Training input} setting applies to both datasets. PointMamba
implementation details are provided in Sec.~\ref{sec:exp_setup}.

On ScanNet, the full-scene-trained PTv3~\citep{wu2024ptv3} checkpoint drops
from 76.6 mIoU under full-scene evaluation to 62.1 with chunked inputs.
The full-scene score is a reference obtained with a larger per-forward
point budget.
Training directly on chunks recovers only part of this loss.
PoinTTT carries a compact recurrent state across chunks and outperforms both
PTv3 training variants and PointMamba~\citep{liang2024pointmamba} under the
same chunked evaluation. On nuScenes, temporal adaptation likewise improves
over PTv3 under the single-pass protocol (Tab.~\ref{tab:pilot}).

Fig.~\ref{fig:teaser}c shows that the memory usage of PTv3 and PointMamba
grows with input size, while PoinTTT keeps activation memory bounded with
fixed-size chunks. These pilot results motivate combining local point-cloud
encoding with persistent scene memory.

\section{Related Work}

\subsection{Point-Cloud Segmentation Backbones}
Point-cloud segmentation has progressed from permutation-invariant set encoders
to architectures with increasingly rich local geometry and context.
PointNet~\citep{qi2017pointnet} and
PointNet++~\citep{qi2017pointnetplusplus} introduced direct neural processing
of point sets, while DGCNN~\citep{wang2019dgcnn} and
KPConv~\citep{thomas2019kpconv} improved local interaction modeling.
MinkowskiNet~\citep{choy20194d} made large voxelized scenes practical, and
Point Transformer~\citep{zhao2021pointtransformer},
PTv2~\citep{wu2022pointtransformerv2}, Swin3D~\citep{yang2025swin3d} and PTv3~\citep{wu2024ptv3}
introduced attention-based architectures for 3D segmentation. Recent systems
further scale this line through point prompt training in
PPT~\citep{wu2024ppt}, multi-dataset pretraining in Point-MoE~\citep{chen2025point}, lighter transformer design in
LitePT~\citep{yue2025litept}, and vanilla transformer variants in
VOLT~\citep{yilmaz2026volt}. PointMamba~\citep{liang2024pointmamba}
explores state-space sequence modeling for point-cloud analysis.
These models provide strong backbones for bounded point sets, voxel sets,
windows, or scenes.
PoinTTT complements this progress by retaining the backbone as a local point-cloud encoder while introducing a compact recurrent state that carries information across chunk boundaries. This allows computation to extend beyond individual chunks and supports spatially or temporally extended inputs.

\subsection{Large-Scale Point-Cloud Understanding}

Large-scale point-cloud benchmarks make the bounded-input assumption visible. Indoor datasets such as S3DIS~\citep{armeni20163d}, Matterport3D~\citep{chang2017matterport3d}, ScanNet~\citep{dai2017scannet}, and ScanNet++~\citep{yeshwanth2023scannetpp}, together with outdoor LiDAR datasets such as Semantic3D~\citep{hackel2017semantic3d}, SemanticKITTI~\citep{behley2019semantickitti}, nuScenes~\citep{caesar2020nuscenes}, Waymo~\citep{sun2020waymo}, DALES~\citep{varney2020dales}, and FOR-instanceV2~\citep{xiang2025forestformer3d}, span diverse environments. Yet mainstream backbones, including point-based methods~\citep{qi2017pointnet,qi2017pointnetplusplus}, voxel and sparse convolutional methods~\citep{graham20183d,chen2022focal,li2025pointcnn++,shi2022pillarnet}, and transformer-based methods~\citep{zhao2021pointtransformer,wu2024ptv3,wu2022pointtransformerv2,yang2025swin3d}, are in practice trained and evaluated on sampled blocks, sliding windows or tiles.

Dynamic point cloud datasets such as MSR-Action3D~\citep{Li2010action}, NTU RGB+D~\citep{shahroudy2016ntu}, Synthia 4D~\citep{choy20194d}, and HOI4D~\citep{liu2022hoi4d}, together with the temporal extensions of outdoor LiDAR benchmarks above, add a temporal version of the same scaling issue. A line of work models point-cloud sequences via spatiotemporal neighborhoods, 4D convolutions, transformer attention, or primitive-level hierarchies~\citep{choy20194d,liu2019meteornet,fan2019pointrnn,fan2021pstnet,wei2022pst2,fan2021p4transformer,wen2022pptr,liu2022hoi4d,liu2025pointnet4d,li2026sts}, but the common protocols still operate on short clips or reset state at clip boundaries. PoinTTT uses a shared recurrent chunk interface for large spatial scenes and full temporal sequences, carrying state across chunk boundaries.

\subsection{State Recurrence and Test-Time Training}

State recurrence has re-emerged as a scalable alternative to explicit all-pair attention. Linear attention~\citep{katharopoulos2020linear,yang2024gated,team2025kimi,yang2024parallelizing,schlag2021linear,schlaglearning} can be viewed as a recurrent accumulation of key-value statistics, state-space models such as Mamba~\citep{gu2023mamba,dao2024transformers,lahotimamba} use compact causal state, and TTT layers~\citep{sun2025ttt} make the state an inner-loop-updated network. LaCT~\citep{zhang2025tttright} improves the efficiency and capacity of this view by updating large chunks with expressive fast-weight MLPs. In 3D, closely related reconstruction systems use TTT or LaCT-like states to compress long image sequences into compact scene representations, including tttLRM~\citep{wang2026tttlrm}, ZipMap~\citep{jin2026zipmap}, and Fast Spatial Memory~\citep{ma2026fsm}. A separate line of work brings test-time learning to point clouds for distribution shift rather than long-range memory: MATE~\citep{mirza2023mate} uses masked autoencoding as an online test-time objective for point cloud classification, SMART-PC~\citep{bahri2025smartpc} adapts robust skeletal representations under point-cloud corruptions, and PointMAC~\citep{jiang2025pointmac} uses meta-learned auxiliary objectives for test-time point-cloud completion. These methods adapt a model to a sample or target distribution, whereas PoinTTT uses fast weights that store context across chunks of the same large scene or temporal sequence, and its update-and-apply recipe applies for cross-chunk recurrence while respecting the unordered nature of points within each chunk.

\renewcommand{\displayequationstyle}{\compactmethoddisplays}
\section{Method}\label{sec:method}

PoinTTT processes arbitrary-extent point clouds as sequences of bounded chunks.
Its core mechanism is a recurrent fast-weight state, implemented with residual
LaCT~\citep{zhang2025tttright} layers, that is updated by each chunk and reused
by subsequent chunks. The underlying point-cloud backbone provides local feature
extraction within each chunk, while LaCT carries information across chunks.
This separates local geometry processing from global context accumulation,
allowing one model to handle both large static scenes and streaming 4D inputs
while keeping activation memory bounded by the size of each chunk.

\begin{figure*}[t]
  \centering
  \includegraphics[width=\linewidth]{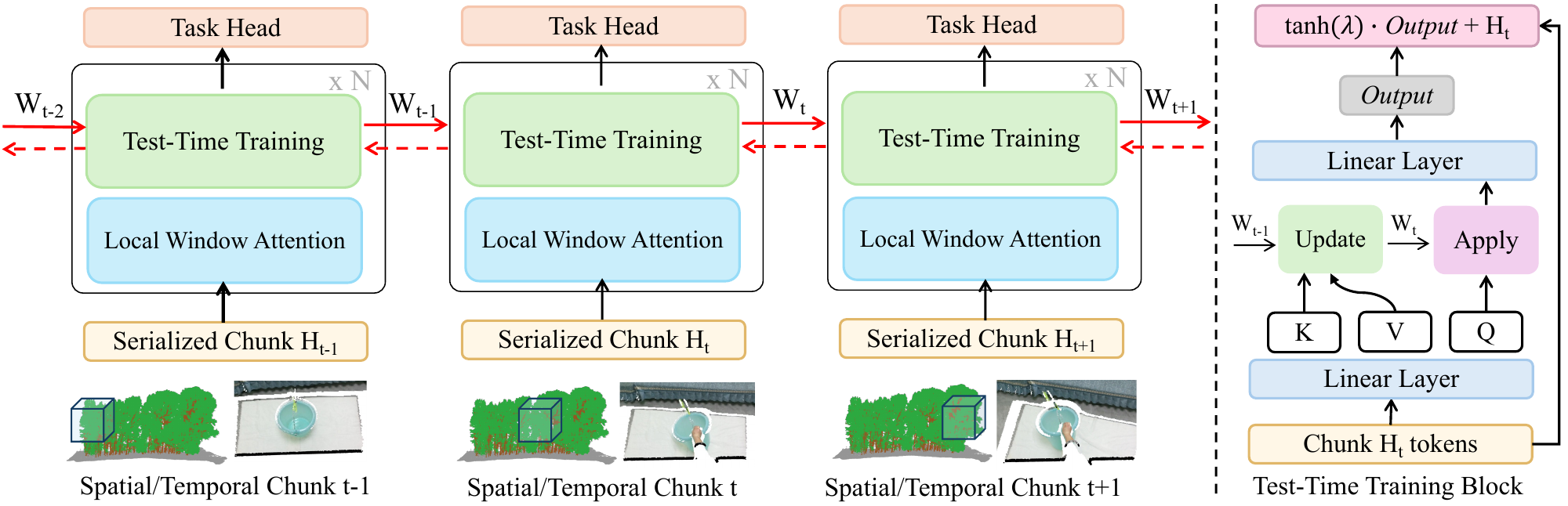}
    \caption{\textbf{PoinTTT uses recurrent fast weights for unified spatial and temporal point-cloud inference.}
    Each step processes one spatial or temporal chunk with a shared
    PTv3-based~\citep{wu2024ptv3} backbone augmented by a residual
    LaCT~\citep{zhang2025tttright} block. LaCT maps chunk features $H_t$ to
    $Q,K,V$, updates fast weights from $W_{t-1}$ to $W_t$ using $K,V$, and applies
    the updated weights to $Q$. The readout is normalized and linearly projected
    before being added to $H_t$ through the gate $\tanh(\lambda)$. Solid arrows denote
    causal propagation for temporal streams, while dashed reverse arrows indicate
    bidirectional propagation over spatial chunks. The resulting compact recurrent
    state enables large-scene and streaming 4D inference without storing full-scene
    activations.}
  \label{fig:method}
\end{figure*}

\subsection{Chunked Point-Cloud Segmentation}
\label{sec:protocol}

PoinTTT is designed around a chunked inference interface: the input may be
arbitrarily large, but the model only receives a bounded point-cloud chunk at
each step. Let a semantic segmentation scene be
$\mathcal{S}=\{(p_i,a_i,y_i)\}_{i=1}^{n}$, where $n$ is the number of points, $p_i\in\mathbb{R}^{3}$ is a
point coordinate, $a_i$ denotes optional point attributes, and
$y_i$ is the semantic label. At inference time, the scene is
represented as an ordered chunk sequence
$\Pi(\mathcal{S})=(\mathcal{P}_1,\ldots,\mathcal{P}_T)$, where each chunk is
$\mathcal{P}_t=\{(p_{t,j},a_{t,j})\}_{j=1}^{M_t}$ with $M_t\le M$. Here $M$ is
the fixed chunk budget and $T$ grows with the spatial or temporal extent of the
observation. The goal is to predict one semantic label for every point while
keeping activation memory bounded by $M$, rather than by the full scene size.

Chunk ordering depends on the acquisition setting. For static point clouds, we
voxelize the scene on a fixed grid, encode each voxel's integer coordinate with
a 3D Hilbert curve, sort the resulting indices, and partition the 1D ordering
into consecutive fixed-size blocks. Hilbert locality keeps each chunk spatially
compact. For 4D point clouds, each short temporal segment of frames is voxelized
and forms one chunk. In both cases, the fast-weight state evolves along the
chunk axis. This shared recurrent formulation supports offline full-scene
segmentation and online segmentation of successive point-cloud frames.

\subsection{LaCT as Recurrent Scene State}
\label{sec:lact}

A LaCT~\citep{zhang2025tttright} layer stores scene context in the
\emph{fast weights} of a small neural network. Each chunk updates these weights
through a key--value association task, then queries the updated network to
retrieve context. We use a SwiGLU MLP~\citep{shazeer2020glu} with parameters
$W=\{W^{(0)},W^{(1)},W^{(2)}\}$ and mapping
$f_W(x)=W^{(1)}(\mathrm{SiLU}(W^{(0)}x)\odot W^{(2)}x)$.
Superscripts distinguish the three matrices, while the subscript in $W_t$
identifies the current recurrent step.

Given backbone features $H_t\in\mathbb{R}^{M_t\times d}$, we first apply layer
normalization, multiply by the projection matrix $W^{qkv}$, and apply SiLU.
Splitting the result yields queries $\widetilde{\mathbf Q}_t$, keys
$\widetilde{\mathbf K}_t$, and values $\mathbf V_t$. We then $\ell_2$-normalize
each query and key vector using $z/(\|z\|_2+\epsilon)$ to obtain
$\mathbf Q_t$ and $\mathbf K_t$, respectively. Keys and values supply the
update, while queries read from the updated state.

The update uses a chunk-local associative TTT objective,
\begin{equation*}
  \mathcal{L}_{\mathrm{TTT}}(W;H_t)
  =
  -
  \sum_{j=1}^{M_t}
  \left\langle
  f_W(\mathbf K_{t,j}),
  \mathbf V_{t,j}
  \right\rangle ,
\end{equation*}
which encourages the fast-weight network to associate each key with its value.
Starting from the incoming state $W_{t-1}$, we take a gradient step for each
matrix $W^{(r)}$, $r\in\{0,1,2\}$. The step uses Newton--Schulz
orthogonalization, denoted by $\mathrm{NS}$, followed by $\mathrm{Norm}$ to
restore the row-wise weight norm within each head. These operations define
$W_t=\mathrm{Update}(W_{t-1},H_t)$:
\begin{equation}
  \begin{aligned}
  \Delta W_t^{(r)}
    &= -\mathrm{NS}\!\left(
      \nabla_{W^{(r)}}\mathcal{L}_{\mathrm{TTT}}(W_{t-1};H_t)
      \right),\\
  W_t^{(r)}
    &= \mathrm{Norm}\!\left(W_{t-1}^{(r)}+\Delta W_t^{(r)}\right).
  \end{aligned}
  \label{eq:lact_update}
\end{equation}

We query the updated state and project its output:
\begin{equation}
  O_t
  =
  \mathrm{Apply}(W_t,H_t)
  =
  P_{\mathrm{out}}\!\left(f_{W_t}(\mathbf Q_t)\right).
  \label{eq:lact_apply}
\end{equation}
Here $P_{\mathrm{out}}$ applies per-head layer normalization, concatenates
the heads, and performs the learned output projection shown as the Linear
Layer after Apply in Fig.~\ref{fig:method}.
Thus, $W_t$ serves two roles: it conditions the current chunk output through
the query features, and it is carried forward as a compact recurrent scene
memory that accumulates context as new chunks arrive.

\subsection{PoinTTT Architecture}
\label{sec:pointtt}

PoinTTT augments a point-cloud backbone with residual
LaCT~\citep{zhang2025tttright} layers. The backbone extracts local geometric
features inside each chunk, while LaCT carries information across chunks
through the recurrent fast-weight state in Eq.~\eqref{eq:lact_update}. We
insert LaCT layers in both the encoder and decoder, allowing both geometric and
semantic features to accumulate cross-chunk context.

Each LaCT layer is inserted as a gated residual branch,
\begin{equation*}
  H_t^{\mathrm{out}} = H_t + \tanh(\lambda)\,O_t.
\end{equation*}
Here $O_t$ is the projected LaCT output from Eq.~\eqref{eq:lact_apply}, $H_t$ denotes the
backbone features for chunk $\mathcal{P}_t$, and $\lambda$ is initialized to
zero. Thus, PoinTTT starts from the original backbone behavior and learns to add
cross-chunk computation during training. Spatial and temporal inference use
the same residual LaCT block, with fast weights propagated according to the
chosen inference mode.

The recurrent interface has two modes, depending on whether all chunks are
available before prediction.

\paragraph{Offline spatial inference.}
For static scenes, all chunks are available before prediction. PoinTTT
first scans the chunks in forward order under \texttt{no\_grad}, using
Eq.~\eqref{eq:lact_update}:
\begin{equation*}
  W_t = \mathrm{Update}(W_{t-1},H_t),
  \qquad W_0 = W_{\mathrm{init}}.
\end{equation*}
This pass initializes the fast weights with full-scene context without storing
full-scene activations.

PoinTTT then runs an \emph{apply} pass over the sequence in reverse order,
starting from the final state $W_T$, and produces the chunk features
$\widetilde{H}_t$ and logits $\hat{Y}_t$,
\begin{equation*}
  \begin{aligned}
  \widetilde{H}_t
  &=
  H_t+\tanh(\lambda)\,\mathrm{Apply}(W_T,H_t),\\
  \hat{Y}_t
  &=
  \mathrm{Decode}(\widetilde{H}_t).
  \end{aligned}
\end{equation*}
The reverse apply pass is differentiable with respect to both the LaCT layers
and the surrounding backbone parameters during training. Gradients are
accumulated across the apply sequence and backpropagated after the full pass
completes. At inference time, this update-and-apply procedure predicts each chunk
using state informed by the full spatial sequence, while keeping activation
memory bounded by chunk size.

\paragraph{Online temporal inference.}

For dynamic scenes, future observations are unavailable when the current chunk
must be predicted. PoinTTT therefore uses causal recurrence:

\begin{equation*}
  \begin{aligned}
  W_t
  &=
  \mathrm{Update}(W_{t-1},H_t),
  \\
  \widetilde{H}_t
  &=
  H_t+\tanh(\lambda)\,\mathrm{Apply}(W_t,H_t),
  \\
  \hat{Y}_t
  &=
  \mathrm{Decode}(\widetilde{H}_t).
  \end{aligned}
\end{equation*}

The prediction at step $t$ depends only on chunks observed up to time $t$,
making the model online in the standard streaming sense. The fast weights act
as recurrent memory over the 4D sequence. During training, gradients are
accumulated across the sequence and backpropagated after the full sequence has
been processed.

\renewcommand{\displayequationstyle}{}
\section{Experiments}
\label{sec:experiments}

\begin{table}[t]
  \centering
  \small
  \caption{Semantic segmentation on the FOR-instanceV2 test split. Best
  results are in \textbf{bold}, second-best are \underline{underlined}.}
  \label{tab:forinstance}
  \begin{tabular*}{\linewidth}{@{\extracolsep{\fill}}lcccc@{}}
    \toprule
    Method & Ground & Wood & Leaf & mIoU \\
    \midrule
        PointMamba~\citep{liang2024pointmamba}       & 95.9               & 16.1               & 91.9               & 68.0               \\
    ForAINetV2\_R8~\citep{xiang2024forainet}       & 98.3 & 68.1 & 94.3 & 86.9 \\
    ForAINetV2\_R16~\citep{xiang2024forainet}      & 98.8    & \underline{69.2}    & 94.5    & \underline{87.5}    \\
    OneFormer3D~\citep{kolodiazhnyi2024oneformer3d}          & 97.8             & 63.1             & 93.4             & 84.8             \\
    ForestFormer3D~\citep{xiang2025forestformer3d}       & 96.9             & 67.7             & 94.1             & 86.2             \\
    \midrule
    PTv3~\citep{wu2024ptv3}       & \underline{99.0}               & 65.9               & \underline{94.7}               & 86.5               \\
    PoinTTT (ours)       & \textbf{99.3}               & \textbf{69.3}               & \textbf{95.1}             & \textbf{87.9}              \\
    \bottomrule
  \end{tabular*}
\end{table}

We evaluate whether PoinTTT scales beyond the ScanNet and nuScenes pilot
study (Sec.~\ref{sec:pilot}) to arbitrary-extent spatial and temporal inputs.
We study two settings:
\textit{forest segmentation}, where the number of spatial chunks grows with scan size,
and \textit{online 4D segmentation}, where chunks arrive over time. Unlike standard
scan- or clip-based inference, PoinTTT initializes its fast-weight state once
per test scan or video and processes the entire input recurrently before
computing the final metric.

\subsection{Setup}
\label{sec:exp_setup}
\noindent \textbf{Datasets and metrics.}
FOR-instanceV2~\citep{xiang2025forestformer3d} contains large forest LiDAR
plots with ground, wood, and leaf labels; individual scenes range from millions
of points to roughly $4\!\times\!10^{8}$ points. We report class IoU and mIoU.
HOI4D~\citep{liu2022hoi4d} contains dynamic hand--object point-cloud sequences;
we report semantic segmentation mIoU over the temporal stream.

\noindent \textbf{Training.}
For FOR-instanceV2 and HOI4D, we first pretrain the PTv3~\citep{wu2024ptv3}
backbone on fixed-size point-cloud chunks using the standard point-wise
cross-entropy segmentation loss. This stage learns local geometric and
semantic features for the chunk encoder. We then insert residual
LaCT~\citep{zhang2025tttright} layers and finetune PoinTTT end-to-end,
updating both the backbone and recurrent layers with the update-and-apply
recurrence active. Forest pretraining and finetuning use 100 and 200 epochs,
respectively; HOI4D uses 50 epochs for each stage. Both stages use AdamW with
OneCycleLR schedules. Forest training uses a single A100 GPU in both stages.
HOI4D pretraining runs on one A100 GPU, while finetuning uses eight A100 GPUs.
Inference for both benchmarks is performed on a single A100 GPU.
See the appendix for batch sizes, parameter-group learning rates, and
scheduler settings for pretraining and finetuning.

\noindent \textbf{Mamba baselines.}
For spatial modeling, we adapt
PointMamba~\citep{liang2024pointmamba} to scene-level semantic segmentation
using the same PTv3-style scene processing protocol as our baselines, while
preserving its original tokenization and serialization. For the PTv3+Mamba
control in Sec.~\ref{sec:exp_hoi4d}, we use the temporal Mamba implementation
from PointNet4D~\citep{liu2025pointnet4d}.

\noindent \textbf{Evaluation pipeline.} We follow the chunked point-cloud formulation in Sec.~\ref{sec:protocol}.
For forest segmentation, non-recurrent baselines process local crops
independently and merge their point-wise predictions back into the full plot.
This gives them access to the same input points and evaluation labels, but not
to persistent information across chunks. PoinTTT instead runs offline spatial
recurrence over all chunks in the plot: an update pass traverses the chunk
sequence to build a fast-weight state, and an apply pass decodes per-chunk
predictions using that state; each chunk has at most
102{,}400 points. Predictions are then reassembled
in the original plot coordinates before computing IoU over the full point set.

For HOI4D, published semantic segmentation protocols use local temporal
windows. PoinTTT runs online recurrence over the full
sequence: the state is initialized once per video, updated by each three-frame
chunk, and carried forward without observing future chunks. All methods use
the same semantic labels and evaluation metrics. PoinTTT carries recurrent
context across the full video instead of treating short clips as independent
inputs.

\begin{table}[t]
  \centering
  \small
  \caption{Semantic segmentation on HOI4D. $\dagger$ marks published
  offline supervised results from PointNet4D~\citep{liu2025pointnet4d},
  Table~9; P4Mamba is its online result from Table~7.
  PTv3 uses fixed windows;
  PTv3+Mamba and PoinTTT share the PTv3 backbone and full-video streaming protocol.
  }
  \label{tab:hoi4d}
  \label{tab:ablate_memory}
  \setlength{\tabcolsep}{3pt}
  \begin{tabular*}{\linewidth}{@{\extracolsep{\fill}}lccc@{}}
    \toprule
    Method & Input window & Context horizon & mIoU \\
    \midrule
    P4Mamba~\cite{liu2025pointnet4d}  & 3 & 3 & 39.8 \\
    P4Transformer$^\dagger$~\citep{fan2021p4transformer} & 3 & 3 & 40.1 \\
    PPTr$^\dagger$~\citep{wen2022pptr}                   & 3 & 3 & 41.0 \\
    PointNet4D$^\dagger$~\citep{liu2025pointnet4d}       & 3 & 3 & 41.6 \\
    PointNet4D++$^\dagger$~\citep{liu2025pointnet4d}     & 3 & 3 & 41.9 \\
    \midrule
    PTv3 (our baseline) & 3 & 3 & 41.9 \\
    PTv3+Mamba & 3 & full video (300) & 39.9 \\
    PoinTTT (ours) & 3 & full video (300) & \textbf{42.6} \\
    \bottomrule
  \end{tabular*}
\end{table}

\begin{figure*}[t]
  \centering
  \includegraphics[width=\linewidth]{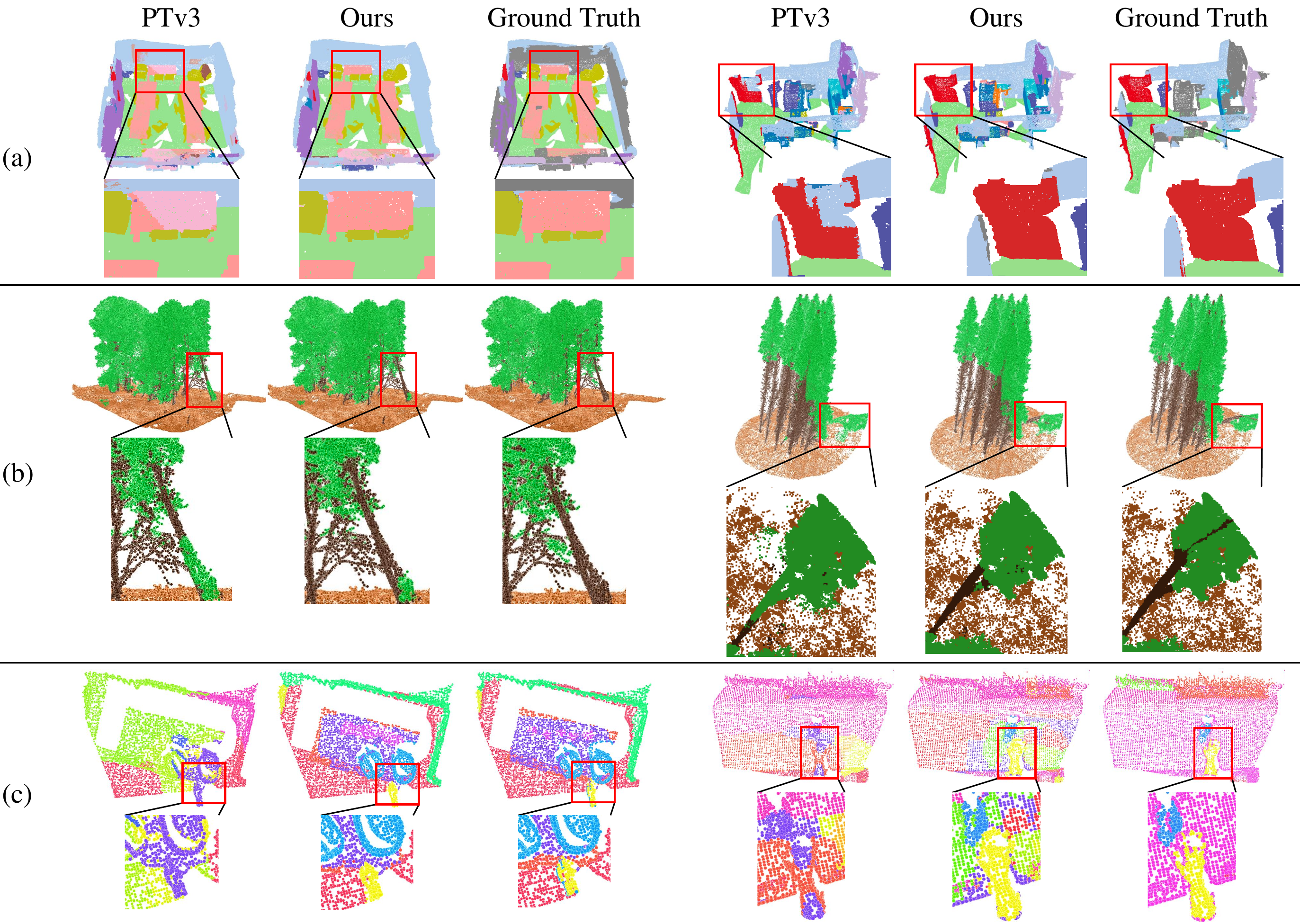}
  \caption{\textbf{Qualitative comparison between PoinTTT and PTv3.}
Two examples per dataset show semantic predictions and ground truth; red boxes
mark enlarged regions. PoinTTT (Ours) reduces chunk-boundary artifacts on
(a) ScanNet~\citep{dai2017scannet}, better preserves tilted trunks and irregular
branches on (b) FOR-instanceV2~\citep{xiang2025forestformer3d}, and more clearly
separates moving objects from the background on (c) HOI4D~\citep{liu2022hoi4d}.}
  \label{fig:qual}
\end{figure*}

\begin{table*}[t]
  \centering
  \small
  \renewcommand{\arraystretch}{1.0}
  \setlength{\tabcolsep}{2pt}
  \newcommand{\ablationrowstrut}{\rule[-3.72pt]{0pt}{12.4pt}}
  \newcommand{\ablationtworowstrut}{\rule[-5.58pt]{0pt}{18.6pt}}
  \newcommand{\ablationupdaterowstrut}{\rule[-4.96pt]{0pt}{16.53333pt}}
  \newcommand{\memoryrowstrut}{\rule[-7.44pt]{0pt}{24.8pt}}
  \caption{Ablations of recurrent aggregation and model design. Metrics are
  percentages except (i). Panels (a)--(g) use
  FOR-instanceV2; (h)--(i) use ScanNet. Panel (a) disables aggregation in
  training and evaluation; (b) matches total training epochs.
  In (d)--(g), $\dagger$ marks the default. Random membership removes
  within-chunk spatial coherence. Panel (i) compares full-scene PTv3 with
  chunked PoinTTT.}
  \label{tab:ablations}

  \begin{minipage}[t]{.32\linewidth}
    \vspace{0pt}
    \centering
    \textbf{\strut(a) Aggregation}\par\vspace{2pt}
    \label{tab:ablate_aggregation}
    \begin{tabular*}{\linewidth}{@{\extracolsep{\fill}}cl@{\extracolsep{0pt}\hspace{7pt}}c@{\extracolsep{\fill}}c@{}}
      \toprule
      & PoinTTT variant & mIoU & \\
      \midrule
      & \ablationtworowstrut No aggregation & 86.8 & \\
      & \ablationtworowstrut Full model & 87.9 & \\
      \bottomrule
    \end{tabular*}
  \end{minipage}\hfill
  \begin{minipage}[t]{.32\linewidth}
    \vspace{0pt}
    \centering
    \textbf{\strut(b) Training budget}\par\vspace{2pt}
    \label{tab:ablate_train}
    \begin{tabular*}{\linewidth}{@{\extracolsep{\fill}}cl@{\extracolsep{0pt}\hspace{7pt}}c@{\hspace{7pt}}c@{\hspace{7pt}}c@{\extracolsep{\fill}}c@{}}
      \toprule
      & Method & mIoU & mAcc & allAcc & \\
      \midrule
      & \ablationtworowstrut PTv3 & 86.8 & 90.7 & 95.7 & \\
      & \ablationtworowstrut PoinTTT & \textbf{87.9} & \textbf{91.9} & \textbf{96.0} & \\
      \bottomrule
    \end{tabular*}
  \end{minipage}\hfill
  \begin{minipage}[t]{.32\linewidth}
    \vspace{0pt}
    \centering
    \textbf{\strut(c) LaCT placement}\par\vspace{2pt}
    \label{tab:ablate_position}
    \begin{tabular*}{\linewidth}{@{\extracolsep{\fill}}cl@{\extracolsep{0pt}\hspace{7pt}}c@{\hspace{7pt}}c@{\hspace{7pt}}c@{\extracolsep{\fill}}c@{}}
      \toprule
      & Placement & mIoU & mAcc & allAcc & \\
      \midrule
      & \ablationrowstrut encoder & \underline{87.3} & \textbf{91.9} & \underline{95.8} & \\
      & \ablationrowstrut decoder & 87.1 & \underline{90.6} & \underline{95.8} & \\
      & \ablationrowstrut all & \textbf{87.9} & \textbf{91.9} & \textbf{96.0} & \\
      \bottomrule
    \end{tabular*}
  \end{minipage}

  \par\vspace{5pt}
  \begin{minipage}[t]{.32\linewidth}
    \vspace{0pt}
    \centering
    \textbf{\strut(d) Chunk size}\par\vspace{2pt}
    \label{tab:rebuttal_design}
    \label{tab:ablate_chunk_size}
    \begin{tabular*}{\linewidth}{@{\extracolsep{\fill}}cl@{\extracolsep{0pt}\hspace{6pt}}c@{\extracolsep{\fill}}c@{}}
      \toprule
      & Points/chunk & mIoU & \\
      \midrule
      & \ablationrowstrut 51,200 & 87.44 & \\
      & \ablationrowstrut 102,400$^\dagger$ & 87.89 & \\
      & \ablationrowstrut 204,800 & 87.68 & \\
      \bottomrule
    \end{tabular*}
  \end{minipage}\hfill
  \begin{minipage}[t]{.32\linewidth}
    \vspace{0pt}
    \centering
    \textbf{\strut(e) Chunk construction}\par\vspace{2pt}
    \label{tab:ablate_order}
    \begin{tabular*}{\linewidth}{@{\extracolsep{\fill}}cl@{\extracolsep{0pt}\hspace{6pt}}c@{\extracolsep{\fill}}c@{}}
      \toprule
      & Construction & mIoU & \\
      \midrule
      & \ablationrowstrut Hilbert$^\dagger$ & 87.89 & \\
      & \ablationrowstrut Z-order & 87.56 & \\
      & \ablationrowstrut Random membership & 87.08 & \\
      \bottomrule
    \end{tabular*}
  \end{minipage}\hfill
  \begin{minipage}[t]{.32\linewidth}
    \vspace{0pt}
    \centering
    \textbf{\strut(f) Recurrent depth}\par\vspace{2pt}
    \label{tab:ablate_depth}
    \begin{tabular*}{\linewidth}{@{\extracolsep{\fill}}cc@{\extracolsep{0pt}\hspace{6pt}}c@{\hspace{6pt}}c@{\hspace{6pt}}c@{\extracolsep{\fill}}c@{}}
      \toprule
      & Encoder & Decoder & Layers & mIoU & \\
      \midrule
      & \ablationrowstrut 4 & 0 & 2 & 87.45 & \\
      & \ablationrowstrut 3,4 & 0,1 & 4 & 87.62 & \\
      & \ablationrowstrut 1--4$^\dagger$ & 0--3 & 8 & 87.89 & \\
      \bottomrule
    \end{tabular*}
  \end{minipage}

  \par\vspace{5pt}
  \begin{minipage}[t]{.32\linewidth}
    \vspace{0pt}
    \centering
    \textbf{\strut(g) Update steps}\par\vspace{2pt}
    \label{tab:ablate_updates}
    \begin{tabular*}{\linewidth}{@{\extracolsep{\fill}}cl@{\extracolsep{0pt}\hspace{6pt}}c@{\extracolsep{\fill}}c@{}}
      \toprule
      & Steps/chunk & mIoU & \\
      \midrule
      & \ablationupdaterowstrut 1$^\dagger$ & 87.89 & \\
      & \ablationupdaterowstrut 2 & 87.64 & \\
      & \ablationupdaterowstrut 4 & 87.56 & \\
      \bottomrule
    \end{tabular*}
  \end{minipage}\hfill
  \begin{minipage}[t]{.32\linewidth}
    \vspace{0pt}
    \centering
    \textbf{\strut(h) Point budgets (ScanNet)}\par\vspace{2pt}
    \label{tab:scannet_budget}
  \begin{tabular*}{\linewidth}{@{\extracolsep{\fill}}cr@{\extracolsep{0pt}\hspace{5pt}}c@{\hspace{5pt}}c@{\hspace{5pt}}c@{\extracolsep{\fill}}c@{}}
    \toprule
    & Points/chunk & PTv3 & PoinTTT & Gain & \\
    \midrule
    & \ablationrowstrut 4{,}096 & 58.3 & 59.2 & +0.9 & \\
    & \ablationrowstrut 20{,}480 & 65.9 & 70.2 & +4.3 & \\
    & \ablationrowstrut 40{,}960 & 67.1 & 73.1 & +6.0 & \\
    & \ablationrowstrut 102{,}400 & 68.4 & 74.4 & +6.0 & \\
    \bottomrule
  \end{tabular*}

  \end{minipage}\hfill
  \begin{minipage}[t]{.32\linewidth}
    \vspace{0pt}
    \centering
    \textbf{\strut(i) Memory footprint (ScanNet)}\par\vspace{2pt}
    \label{tab:ablate_memory_scaling}
    \begin{tabular*}{\linewidth}{@{\extracolsep{\fill}}cl@{\extracolsep{0pt}\hspace{8pt}}c@{\hspace{8pt}}c@{\extracolsep{\fill}}c@{}}
      \toprule
      & Memory & PTv3 & PoinTTT & \\
      \midrule
      & \memoryrowstrut Peak VRAM & 19 GB & \textbf{0.3 GB} & \\
      & \memoryrowstrut Fast-weight state & -- & 1 MB & \\
      \bottomrule
    \end{tabular*}
  \end{minipage}
\end{table*}

\subsection{Forest Spatial Segmentation}
\label{sec:exp_forest}

Tab.~\ref{tab:forinstance} compares PoinTTT with prior forest segmentation
methods on FOR-instanceV2~\citep{xiang2025forestformer3d}. PoinTTT is trained
only on fixed-size chunks, but at test time it is applied to each forest plot as
a full-scene recurrent sequence over all chunks. Thus, the model keeps the same
bounded-memory interface used during training while allowing information to be
carried across the entire plot. PoinTTT reaches
87.9 mIoU, improving over PTv3~\citep{wu2024ptv3} by 1.4 points and over
ForestFormer3D~\citep{xiang2025forestformer3d} by 1.7 points.

The largest gain appears on the \textit{wood} class, where stems and branches often
extend across local partitions. This suggests that full-plot context is useful
for point-cloud understanding, and that fast-weight recurrence provides an
effective memory mechanism for aggregating such cross-chunk information.
Qualitative results in Fig.~\ref{fig:qual}(b) further show that PoinTTT better
preserves connected tree structures during offline spatial inference.

\subsection{Online 4D Temporal Segmentation}
\label{sec:exp_hoi4d}

\noindent \textbf{Streaming inference beyond the local window.}
PoinTTT processes each HOI4D video as a stream of three-frame chunks,
each serving as the \emph{input window} jointly encoded at one step.
The fast-weight state is initialized once per video and carried across chunks;
predictions depend only on observations received so far. The \emph{context
horizon} grows with the observed video prefix, reaching approximately
300 frames at the end of a video. The recurrence has no preset video-length limit:
processing can continue with the same chunk budget and fixed-size state.
The reported experiment tests this capability over complete HOI4D videos.

\noindent \textbf{Published protocols and matched memory mechanisms.}
Prior HOI4D baselines use short-clip protocols, typically three frames,
with ten-frame variants also reported~\citep{liu2025pointnet4d}.
Our PTv3 baseline follows this local-window setup and is trained and
evaluated on three-frame clips. Tab.~\ref{tab:hoi4d} reports the published
three-frame results, distinguishing the offline supervised results in
PointNet4D~\citep{liu2025pointnet4d}, Table~9, from its online P4Mamba
result in Table~7.

To compare recurrent memory mechanisms on the same long sequences, we
construct a PTv3+Mamba baseline by replacing LaCT's fast-weight memory with
Mamba selective state-space layers. We retain the same PTv3 backbone,
three-frame input windows, training setup, parameter budget, and full-video
streaming protocol. Both models carry their respective recurrent states
across the complete video. Tab.~\ref{tab:hoi4d} reports 39.9 mIoU for
PTv3+Mamba and 42.6 for PoinTTT.

Complementary HOI4D approaches include LeaF~\citep{liu2023leaf} and
4DMAP, the frame-wise masked autoregressive pretraining scheme introduced
with PointNet4D~\citep{liu2025pointnet4d}. Our comparison here focuses on
4D point-cloud encoder backbones under supervised semantic segmentation
training.

\subsection{Qualitative Comparison}

On ScanNet (Fig.~\ref{fig:qual}), PTv3 exhibits block-shaped and diagonal artifacts introduced by
Hilbert-order chunk partitioning; PoinTTT reduces these artifacts and produces
smoother semantic boundaries. On FOR-instanceV2, PTv3 struggles with tilted
trunks and irregular branches that deviate from dominant vertical structures,
while PoinTTT better preserves their geometric continuity. On HOI4D, PTv3 is
more prone to frame-local mistakes, such as confusing moving objects with nearby
background regions. Temporal context across frames helps PoinTTT make more
stable predictions and distinguish moving objects from the background.

\subsection{Ablations}
\label{sec:ablations}

\noindent \textbf{Memory and capacity.}
Tab.~\ref{tab:ablations}a retains the LaCT modules while disabling
cross-chunk aggregation during both training and evaluation. The added
transformations remain available, but their information is confined to
individual chunks. This model learns under the same restricted context that
it encounters at inference. Its lower accuracy supports using the fast
weights to communicate across chunks, beyond their role in transforming local
features. Tab.~\ref{tab:ablations}b examines the training schedule: extending PTv3 training
to the same total number of epochs still leaves an accuracy gap. Together,
these comparisons motivate making scene context available to local predictions
through a persistent state.

\noindent \textbf{Architecture design.}
Joint encoder--decoder insertion outperforms either placement alone (Tab.~\ref{tab:ablations}c),
suggesting that scene context benefits both feature extraction and point-wise
reconstruction. Each inserted layer maintains its own fast weights and
forms associations from the features at that stage, allowing different
resolutions to maintain different representations of the scene. Extending
recurrence to more encoder and decoder stages improves accuracy (Tab.~\ref{tab:ablations}f),
supporting our choice to distribute scene memory across different levels of
the feature hierarchy.

\noindent \textbf{Chunk size.}
Chunk size controls both the geometry jointly available to the backbone and
the granularity of recurrent updates. For a fixed scene, larger chunks produce
fewer state updates, each incorporating more points; the fast-weight state
capacity remains unchanged. The forest sweep in Tab.~\ref{tab:ablations}d is non-monotonic: the
intermediate budget performs best, and increasing it further does not improve
accuracy. One possible explanation is a balance between local geometric
support and the frequency of context updates across the scene. We use
102{,}400 points per chunk as the best-performing tested setting, processing
fewer points at once than the largest tested budget.

\noindent \textbf{Chunk construction.}
Hilbert and Z-order both outperform random chunk membership (Tab.~\ref{tab:ablations}e). Both curves
keep spatially related points together, providing coherent neighborhoods for
the local point encoder. Random membership disperses these observations across
chunks, changing the geometry available within each forward pass. Recurrent
context can connect the resulting features across chunks, but it does not
give the encoder explicit access to the other chunks' point neighborhoods.
The gap between Hilbert and Z-order is smaller than either curve's gap to
random membership, consistent with the value of retaining spatial coherence
alongside recurrent context. We use Hilbert serialization for the remaining
experiments.

\noindent \textbf{Update steps.}
Each inner update writes key--value associations from the current chunk into
the fast weights. Repeating the update revisits the same observations;
it does not introduce additional points or extend the observed scene.
More updates per chunk do not improve segmentation accuracy (Tab.~\ref{tab:ablations}g).
The associative update objective and the segmentation objective serve
different purposes, so more inner optimization need not yield better
predictions. This also differs from adding recurrent layers in Tab.~\ref{tab:ablations}f, where
each layer writes and reads features from a different stage of the backbone.
We use one update per chunk, which gives the best result in this sweep
without the additional inner iterations.

\noindent \textbf{Input budgets and memory scaling.}
Tab.~\ref{tab:ablations}h extends the pilot study to test whether the
advantage of recurrence persists across different local input budgets.
On ScanNet, both models benefit from larger evaluation chunks, while PoinTTT
outperforms independently chunked PTv3 at every matched point budget.
Both methods eventually process all points in a scene. However, merging
independently predicted chunks does not let an individual prediction use
information from outside its own chunk. PoinTTT can condition that prediction
on information accumulated in the recurrent state. Its advantage at larger
budgets suggests that local geometric coverage and cross-chunk context offer
complementary benefits: a larger chunk strengthens the local input, while
recurrence supplies information beyond it.

Tab.~\ref{tab:ablations}i compares full-scene PTv3 with chunked PoinTTT on the same scene,
using different per-forward point budgets. Full-scene processing keeps
activations for the entire input, whereas PoinTTT processes a bounded chunk
and retains scene information in its fast weights. These weights have a fixed
size per layer, independent of the total point count, and are not a growing
cache of point-wise features. Compressing
context into this state allows information to persist across chunks without
keeping full-scene activations, accounting for the lower peak VRAM of chunked
execution. The appendix provides the per-forward point budgets and the memory
required to store the fast weights.

\section{Conclusion}

We argue that arbitrary-extent point-cloud segmentation should be
treated as a recurrent inference problem, rather than as independent crop or
clip prediction. Chunking provides the necessary bounded-memory interface, but
our ScanNet~\cite{dai2017scannet} pilot shows that chunking alone creates a context bottleneck:
predictions made from local chunks cannot fully recover scene-level reasoning.

Thus, we introduce PoinTTT to address this bottleneck by carrying compact LaCT~\cite{zhang2025tttright} fast weights across
the chunk sequence. These fast weights act as recurrent scene memory, allowing
the model to accumulate context while keeping activation memory tied to the
chunk size. The same mechanism supports both offline spatial inference over
large static scenes and online temporal inference over 4D streams. Across ScanNet, FOR-instanceV2~\cite{xiang2025forestformer3d}, HOI4D~\cite{liu2022hoi4d}, and nuScenes~\cite{caesar2020nuscenes}, PoinTTT improves segmentation under
fixed-memory inference and demonstrates that fast-weight TTT recurrence is a
practical way to connect local point-cloud chunks into coherent full-scene or
full-sequence predictions.

\section*{Acknowledgments}

We thank MathWorks and Adobe for their research gifts. We acknowledge computational support
from the Research Computing and Data Analytics Center at the University of
Virginia, ACCESS, and the NAIRR Pilot.
These resources include Purdue University's Anvil supercomputer
(NSF award OAC-2005632) and the Delta and DeltaAI systems
(NSF award OAC-2005572).

{
\small
\bibliographystyle{ieeenat_fullname}
\bibliography{references}
}

\clearpage
\appendix
\section{Supplementary Material}
\label{app:supplementary}

\subsection{Training Details}
\label{app:training}

The training procedure and baseline adaptations are described in
Sec.~\ref{sec:exp_setup}; Tab.~\ref{tab:training_config} lists the detailed
configurations.

The ScanNet experiments also use a single A100 GPU for PTv3 pretraining,
PoinTTT finetuning, and inference. We will release the code with the exact
configuration files, checkpoint schedules, and hardware settings.

\begin{table}[!ht]
\centering
\small
\caption{Training configurations for PTv3 pretraining and PoinTTT finetuning.
All runs use AdamW and OneCycleLR with \texttt{div\_factor}${=}10$.}
\label{tab:training_config}
\setlength{\tabcolsep}{3pt}
\begin{tabular*}{\columnwidth}{@{\extracolsep{\fill}}lccc@{}}
\toprule
Setting & HOI4D & Forest & ScanNet \\
\midrule
\multicolumn{4}{c}{\textbf{PTv3 Pretraining}} \\
\midrule
Batch size & 64 & 1 & 12 \\
GPUs (A100) & 1 & 1 & 1 \\
Base LR & $6\mathrm{e}{-3}$ & $6\mathrm{e}{-4}$ & $6\mathrm{e}{-3}$ \\
Block LR & $6\mathrm{e}{-4}$ & $6\mathrm{e}{-5}$ & $6\mathrm{e}{-4}$ \\
\texttt{pct\_start} & 5\% & 5\% & 5\% \\
\texttt{final\_div\_factor} & 1000 & 1000 & 1000 \\
\midrule
\multicolumn{4}{c}{\textbf{PoinTTT Finetuning}} \\
\midrule
Batch size & 32 & 1 & 12 \\
GPUs (A100) & 8 & 1 & 1 \\
LR (single group) & $1\mathrm{e}{-4}$ & $1\mathrm{e}{-4}$ & $1\mathrm{e}{-4}$ \\
\texttt{pct\_start} & 5\% & 10\% & 10\% \\
\texttt{final\_div\_factor} & 1000 & 100 & 100 \\
\bottomrule
\end{tabular*}
\end{table}

\subsection{Evaluation Pipeline Details}
\label{app:evaluation}

For static forest plots, PoinTTT resets the fast-weight state at the beginning
of each test plot. The plot is partitioned into a sequence of fixed-size
chunks, the offline update pass processes all chunks to build a plot-level
state, and the apply pass decodes chunk predictions using that state. Per-chunk
predictions are then mapped back to the original points, and class IoU and mIoU
are computed over the full plot.

For 4D sequences, PoinTTT resets the fast-weight state only at the beginning of
each video. At step $t$, the model receives the current three-frame chunk,
updates the state, emits point-wise predictions, and carries the updated state
to $t{+}1$. Future chunks are not used. This differs from window-based HOI4D
evaluation, where methods typically process short windows and reset the
backbone state at clip boundaries.

\subsection{Additional Evaluation Details}
\label{app:additional_ablation}

For the ScanNet memory measurements in Tab.~\ref{tab:ablations}i, PTv3
processes the full scene (about 5.2 million points) in one forward pass.
PoinTTT processes up to $10{,}240$ points per pass. Its recurrent state
holds about $280{,}000$ fast weights and occupies approximately 1~MB.

The published PTv3 score of 80.4 mIoU on nuScenes uses test-time
augmentation. Our nuScenes results in Tab.~\ref{tab:pilot} use a single
pass without augmentation.

\subsection{Limitations and Future Work}
\label{app:limitations}

PoinTTT provides a memory-bounded interface for arbitrary-extent point-cloud
segmentation, and several directions remain for future work. First, because the
fast-weight state is propagated recurrently, inference is naturally sequential
along the chunk axis. This keeps activation memory bounded by chunk size, but
the total runtime still grows with the number of chunks. Future work could
improve inference efficiency with parallel, hierarchical, or cached state
propagation while preserving the recurrent context mechanism.

Second, our experiments cover four complementary settings: ScanNet for indoor
segmentation, FOR-instanceV2 for large forest plots, and HOI4D for online 4D
segmentation, together with nuScenes for outdoor driving LiDAR. Due to compute constraints, we do not scale the evaluation to a
larger collection of datasets or longer training schedules. Extending PoinTTT
to more domains, sensors, and scene types would further test the generality of
fast-weight recurrence for point-cloud understanding.

Finally, PoinTTT is trained in a supervised setting and does not yet use
large-scale point-cloud pretraining. We expect pretraining on diverse 3D and 4D
data to improve robustness, cross-domain transfer, and long-range context
aggregation. This is a natural next step, especially for settings with domain
shift such as forests captured under different sensors, acquisition conditions,
or vegetation distributions. We leave this as future work.

\clearpage

\end{document}